\documentclass[10pt,journal,compsoc]{IEEEtran}

\ifCLASSOPTIONcompsoc
  \usepackage[nocompress]{cite}
\else
  \usepackage{cite}
\fi
\usepackage{cite}
\usepackage{amsmath,amssymb,amsfonts}
\usepackage{algorithmic}
\usepackage{graphicx}
\usepackage{textcomp}
\usepackage{xcolor}
\usepackage{booktabs}
\usepackage[flushleft]{threeparttable}
\usepackage{amsfonts,bm}
\usepackage{multirow}
\usepackage{subfigure}
\usepackage{url}
\usepackage{hyperref}
\usepackage{color}
\usepackage[normalem]{ulem}
\def\bs{\boldsymbol}
\def\BibTeX{{\rm B\kern-.05em{\sc i\kern-.025em b}\kern-.08em
    T\kern-.1667em\lower.7ex\hbox{E}\kern-.125emX}}

\begin{document}
%
\title{Sparse Graph Attention Networks}
%
%
%
%

\author{Yang~Ye, and Shihao~Ji, \IEEEmembership{Senior Member, IEEE}
\IEEEcompsocitemizethanks{\IEEEcompsocthanksitem Y. Ye and S. Ji are with the Department of Computer Science, Georgia State University, Atlanta, GA 30303.\protect\\
E-mail: yye10@student.gsu.edu; sji@gsu.edu
}
}

\IEEEtitleabstractindextext{%
\begin{abstract}
Graph Neural Networks (GNNs) have proved to be an effective representation learning framework for graph-structured data, and have achieved state-of-the-art performance on many practical predictive tasks, such as node classification, link prediction and graph classification. Among the variants of GNNs, Graph Attention Networks (GATs) learn to assign dense attention coefficients over all neighbors of a node for feature aggregation, and improve the performance of many graph learning tasks. However, real-world graphs are often very large and noisy, and GATs are prone to overfitting if not regularized properly. Even worse, the local aggregation mechanism of GATs may fail on disassortative graphs, where nodes within local neighborhood provide more noise than useful information for feature aggregation. In this paper, we propose Sparse Graph Attention Networks (SGATs) that learn sparse attention coefficients under an $L_0$-norm regularization, and the learned sparse attentions are then used for all GNN layers, resulting in an edge-sparsified graph. By doing so, we can identify noisy/task-irrelevant edges, and thus perform feature aggregation on most informative neighbors. Extensive experiments on  synthetic and real-world (assortative and disassortative) graph learning benchmarks demonstrate the superior performance of SGATs. In particular, SGATs can remove about 50\%-80\% edges from large assortative graphs, such as PPI and Reddit, while retaining similar classification accuracies. On disassortative graphs, SGATs prune majority of noisy edges and outperform GATs in classification accuracies by significant margins. Furthermore, the removed edges can be interpreted intuitively and quantitatively. To the best of our knowledge, this is the first graph learning algorithm that shows significant redundancies in graphs and edge-sparsified graphs can achieve similar (on assortative graphs) or sometimes higher (on disassortative graphs) predictive performances than original graphs. Our code is available at~\url{https://github.com/Yangyeeee/SGAT}. 
\end{abstract}

}

\maketitle

\IEEEdisplaynontitleabstractindextext

%
\IEEEpeerreviewmaketitle

\ifCLASSOPTIONcompsoc
\IEEEraisesectionheading{\section{Introduction}\label{sec:introduction}}
\else
\section{Introduction}
\label{sec:introduction}
\fi

%
%
%
%
\IEEEPARstart{G}{raph}-structured data is ubiquitous in many real-world systems, such as social networks~\cite{tang2015line}, biological networks~\cite{Zitnik2017}, and citation networks~\cite{senaimag08}, etc. Graphs can capture interactions (i.e., edges) between individual units (i.e., nodes) and encode data from irregular or non-Euclidean domains to facilitate representation learning and data analysis. Many tasks, from link prediction~\cite{vdberg2017graph}, graph classification~\cite{david15finger} to node classification~\cite{yang16semi}, can be naturally performed on graphs, where effective node embeddings that can preserve both node information and graph structure are required. To learn from graph-structured data, typically an encoder function is needed to project high-dimensional node features into a low-dimensional embedding space such that ``semantically" similar nodes are close to each other in the low-dimensional Euclidean space (e.g., by dot product)~\cite{representation17}.

Recently, various Graph Neural Networks (GNNs) have been proposed to learn such embedding functions~\cite{ScaGorTso09,BruZarSzl14,DefBreVan16,kipf2017semi,hamilton2017inductive,representation17,velickovic2018graph,CheLinLi20}. Traditional node embedding methods, such as matrix factorization ~\cite{cao2015grarep,ou2016asymmetric} and random walk~\cite{deepwalk2014,adi16node2vec}, only rely on adjacent matrix (i.e., graph structure) to encode node similarity. Training in an unsupervised way, these methods employ dot product or co-occurrances on short random walks over graphs to measure the similarity between a pair of nodes. Similar to word embeddings~\cite{MikSutChe13,glove14,pword2vec16}, the learned node embeddings from these methods are simple look-up tables. Other approaches exploit both graph structure and node features in a semi-supervised training procedure for node embeddings~\cite{DefBreVan16,kipf2017semi,hamilton2017inductive,velickovic2018graph}. These methods can be classified into two categories based on how they manipulate the adjacent matrix: (1) spectral graph convolution networks~\cite{ScaGorTso09,BruZarSzl14,DefBreVan16}, and (2) neighbor aggregation or message passing algorithms~\cite{kipf2017semi,hamilton2017inductive,velickovic2018graph}. Spectral graph convolution networks transform graphs to the Fourier domain, effectively converting convolutions over the whole graph into element-wise multiplications in the spectral domain. However, once the graph structure changes, the learned embedding functions have to be retrained or finetuned. On the other hand, the neighbor aggregation algorithms treat each node separately and learn feature representation of each node by aggregating (e.g., weighted-sum) over its neighbors' features. Under the assumption that connected nodes should share similar feature representations, these message passing algorithms leverage local feature aggregation to preserve the locality of each node, and is a generalization of classical convolution operation on images to irregular graph-structured data. For both categories of GNN algorithms, they can stack $k$ layers on top of each other and aggregate features from $k$-hop neighbors.

\begin{figure*}[h]
	\begin{center}
		\includegraphics[width=1\linewidth]{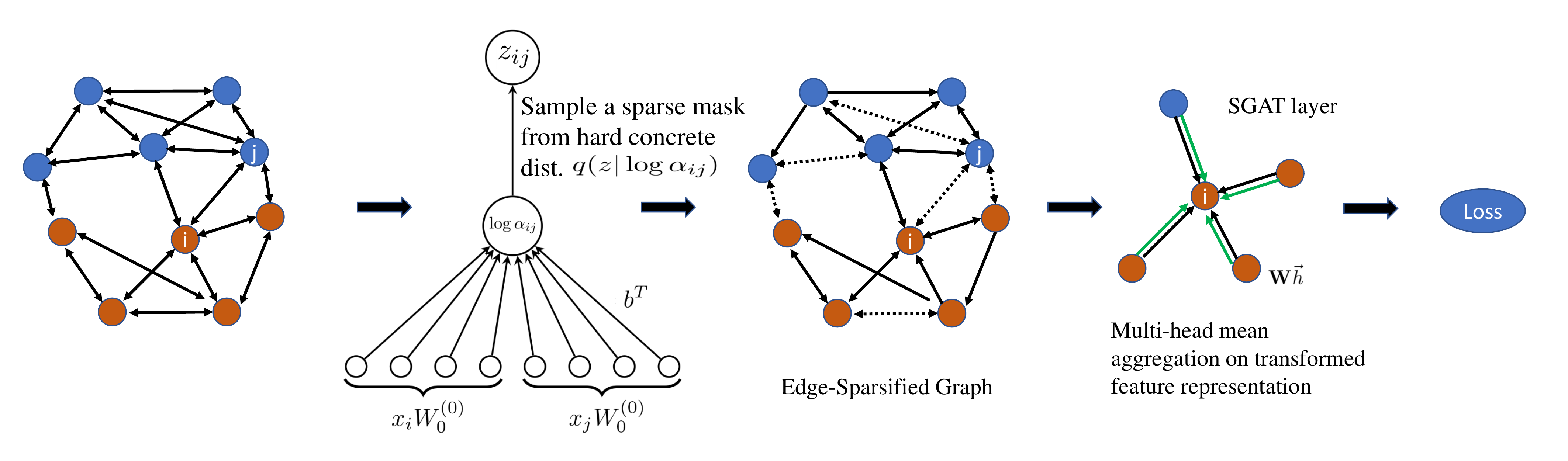}
	\end{center}\vspace{-0.6cm}
	\caption{The overview of SGATs. By attaching a binary mask to each edge, SGATs utilize a sparse attention mechanism (as the output of the mask generator) to guide model to remove noisy/task-irrelevant edges and yield an edge-sparsified graph. In the plot above, the dashed lines denote removed edges. More details are described in Sec.~\ref{sec:method}.}
	\label{fig:sgat}
\end{figure*}

Among all the GNN algorithms, the neighbor aggregation algorithms~\cite{kipf2017semi,hamilton2017inductive,velickovic2018graph} have proved to be more effective and flexible. In particular, Graph Attention Networks (GATs)~\cite{velickovic2018graph} use attention mechanism to calculate edge weights at each layer based on node features, and attend adaptively over all neighbors of a node for representation learning. To increase the expressiveness of the model, GATs further employ multi-head attentions to calculate multiple sets of attention coefficients for aggregation. Although multi-head attentions improve prediction accuracies, our analysis of the learned coefficients shows that multi-head attentions usually learn very similar distributions of attention coefficients (see Sec.~\ref{sec:formulation} for details). This indicates that there might be a significant redundancy in the GAT modeling. In addition, GATs cannot assign an unique attention score for each edge because multiple attention coefficients are generated (from multi-heads) for an edge per layer and the same edge at different layers might receive different attention coefficients. For example, for a 2-layer GAT with 8-head attentions, each edge receives 16 different attention coefficients. The redundancy in the GAT modeling not only adds significant overhead to computation and memory usage but also increases the risk of overfitting. To mitigate these issues, we propose to simplify the architecture of GATs such that only one single attention coefficient is assigned to each edge across all GNN layers. To further reduce the redundancy among edges or remove noisy edges, we incorporate a sparsity constraint into the attention mechanism of GATs. Specifically, we optimize the model under an $L_0$-norm regularization to encourage model to use as fewer edges as possible. As we only employ one attention coefficient for each edge across all GNN layers, what we learn is an \textbf{edge-sparsified} graph with noisy/task-irrelevant edges removed. Our Sparse Graph Attention Networks (SGATs), as shown in Fig.~\ref{fig:sgat}, outperform the original GATs in two aspects: (1) SGATs simplify the architecture of GATs, and this reduces the risk of overfitting, and (2) SGATs can identify noisy/task-irrelevant edges\footnote{We call an edge task-irrelevant or noisy if removing it from graph incurs a similar or improved accuracy for downstream predictive tasks.} of a graph such that an edge-sparsified graph structure can be discovered, which is more robust for downstream classification tasks. As a result, SGAT is a robust graph learning algorithm that can learn from both assortative and disassortative graphs, while GAT fails on disassortative graphs.




\section{Background and Related Work}
In this section, we first introduce our notation and then review prior works related to the neighbor aggregation methods on graphs. Let $G=(V, E)$ denote a graph with a set of nodes $V=\{v_1,\cdots,v_N\}$, connected by a set of edges $E\subseteq V\times V$. Node features are organized in a compact matrix $X\in\mathbb{R}^{N\times D}$ with each row representing the feature vector of one node. Let $A\in\mathbb{R}^{N \times N}$ denote the adjacent matrix that describes graph structure of $G$: $A_{ij}=1$ if there is an edge $e_{ij}$ from node $i$ to node $j$, and 0 otherwise. By adding a self-loop to each node, we have $\tilde{A}=A+I_{N}$ to denote the adjacency matrix of the augmented graph, where $I_N\in\mathbb{R}^{N\times N}$ is an identity matrix. 

For a semi-supervised node classification task, given a set of labeled nodes $\{(v_i, y_i), i=1,\cdots,n\}$, where $y_i$ is the label of node $i$ and $n<N$, we learn a function $f\left(X, A, W\right)$, parameterized by $W$, that takes node features $X$ and graph structure $A$ as inputs and yields a node embedding matrix $H\in\mathbb{R}^{N\times D'}$ for all nodes in $V$; subsequently, $H$ is fed to a classifier to predict the class label of each unlabeled node. To learn the model parameter $W$, we typically minimize an empirical risk over all labeled nodes:
\begin{equation}\label{equ:ce}
\mathcal{R}(W) =\frac{1}{n}\sum_{i=1}^{n}\mathcal{L}\left(f_i(X, A, W), y_{i}\right),
\end{equation}
where $f_i(X,A,W)$ denotes the output of $f(X,A,W)$ for node $i$ and $\mathcal{L}(\cdot)$ is a loss function, such as the cross-entropy loss that measures the compatibility between model predictions and class labels. Although there exist many different GNN algorithms that can solve Eq.~\ref{equ:ce}, the main difference among them is how the encoder function $f(X,A,W)$ is defined.

\subsection{Neighbor Aggregation Methods}
The most effective and flexible graph learning algorithms so far follow a neighbor aggregation mechanism. The basic idea is to learn a parameter-sharing aggregator, which takes feature vector $x_{i}$ of node $i$ and its neighbors' feature vectors $\{x_{j}, j\in\mathcal{N}_i\}$ as inputs and outputs a new feature vector for node $i$. Essentially, the aggregator function aggregates lower-level features of a node and its neighbors and generates high-level feature representations. The popular Graph Convolution Networks (GCNs)~\cite{kipf2017semi} fall into the category of neighbor aggregation. For a 2-layer GCN, its encoder function can be expressed as:
\begin{equation}
    f(X, A, W)=\operatorname{softmax}\left(\hat{A} \sigma(\hat{A} X W^{(0)}) W^{(1)}\right), 
\end{equation} 
where $\hat{A}=\tilde{D}^{-1 / 2} \tilde{A} \tilde{D}^{-1 / 2}$, $\tilde{D}_{i i}=\sum_{j} \tilde{A}_{i j}$, and $W^{(\cdot)}$s are the learnable parameters of GCNs. Apparently, GCNs define the aggregation coefficients as the symmetrically normalized adjacency matrix $\hat{A}$, and these coefficients are shared across all GCN layers. More specifically, the aggregator of GCNs can be expressed as 
\begin{equation}\label{eq:gcn_agg}
  h_i^{(l+1)}=\sigma\left(\sum_{j\in\mathcal{N}_i} \hat{A}_{i j} h_{j}^{(l)} W^{(l)}\right),
\end{equation}
where $h_{j}^{(l)}$ is the hidden representation of node $j$ at layer $l$, $h^{(0)}=X$, and $\mathcal{N}_i$ denotes the set of all the neighbors of node $i$, including itself.

Since a fixed adjacency matrix $\hat{A}$ is used for feature aggregation, GCNs can only be used for the transductive learning tasks, and if the graph structure changes, the whole GCN model needs to be retrained or fine-tuned. To support inductive learning, GraphSage~\cite{hamilton2017inductive} proposes to learn parameterized aggregators (e.g., mean, max-pooling or LSTM aggregator) that can be used for feature aggregation on unseen nodes or graphs. To support large-scale graph learning tasks, GraphSage uniformly samples a fixed number of neighbors per node and performs computation on a sampled sub-graph at each iteration. Although it can reduce computational cost and memory usage significantly, its accuracies suffer from random sampling and partial neighbor aggregation. 

\subsection{Graph Attention Networks}
Recently, attention networks have achieved state-of-the-art results in many computer vision and natural language processing tasks, such as image captioning~\cite{XuBaKir15} and machine translation~\cite{BahChoBen15}. By attending over a set of inputs, attention mechanism can decide which parts of inputs to attend to in order to gather the most useful information. Extending the attention mechanism to graph-structured data, Graph Attention Networks (GATs)~\cite{velickovic2018graph} utilize an attention-based aggregator to generate attention coefficients over all neighbors of a node for feature aggregation. In particular, the aggregator function of GATs is similar to that of GCNs:
\begin{equation}\label{eq:gat_agg}
    h_{i}^{(l+1)} =\sigma\left(\sum_{j\in\mathcal{N}_i} a_{i j}^{(l)} h_{j}^{(l)}W^{(l)}\right), 
\end{equation}
except that (1) $a_{ij}^{(l)}$ is the attention coefficient of edge $e_{ij}$ at layer $l$, assigned by an attention function other than by a predefined $\hat{A}$, and (2) different layers utilize different attention functions, while GCNs share a predefined $\hat{A}$ across all layers. To increase the capacity of attention mechanism, GATs further exploit multi-head attentions for feature aggregation: each head works independently to aggregate information, and all the outputs of multi-heads are then concatenated to form a new feature representation for the next layer. In principle, the learned attention coefficient can be viewed as an importance score of an edge. However, since each edge receives multiple attention coefficients at a layer and the same edge at a different layer has a different set of attention coefficients, GATs cannot assign an unique importance score to quantify the significance of an edge. 

Built on the basic framework of GATs, our SGATs introduce a sparse attention mechanism via an $L_0$-norm regularization for feature aggregation. Furthermore, we only assign one attention coefficient (or importance score) to each edge across all layers. As a result, we can identify important edges of a graph and remove noisy/task-irrelevant ones while retaining similar or sometimes even higher predictive performances on downstream classification tasks. Our results demonstrate that there is a significant amount of redundancies in graphs (e.g., 50\%-80\% of edges in assortative graphs like PPI and Reddit, and over 88\% edges in disassortative graphs) that can be removed to achieve similar or improved classification accuracies.

\subsection{Graph Sparsification} 
There are some prior works related to SGATs in terms of graph sparsification~\cite{spectrals,spectral1,dropedge,BayesGNN,chen2020label,NeuralSparse20,luo2021learning,SuperGAT21}. Spectral graph sparsification~\cite{spectrals,spectral1} aims to remove unnecessary edges for graph compression. Specifically, it identifies a sparse subgraph whose Laplacian matrix can approximate the original Laplacian matrix well. However, these algorithms do not utilize node representations for graph compression and are not suitable for semi-supervised node classification tasks considered in the paper. DropEdge~\cite{dropedge} (and its Bayesian treatment~\cite{BayesGNN}) propose to stochastically drop edges from graphs to regularize the training of GNNs. Specifically, DropEdge randomly removes a certain number of edges from an input graph at each training iteration to prevent the overfitting and oversmoothing issues~\cite{LiHanWu18}. At validation or test phase, DropEdge is disabled and the full input graph is utilized. This method shares the same spirit of Dropout~\cite{dropout14} and is an intuitive extension to graph structured data. However, DropEdge does not induce an edge-sparsified graph since different subsets of edges are removed at different training iterations and the full graph is utilized for validation and test, while SGAT learns an edge-sparsified graph by removing noisy/task-irrelevant edges permanently from input graphs. Because of these discrepancies, these methods are not directly comparable to SGAT. Recently, Chen et al.~\cite{chen2020label} propose LAGCN to add/remove edges based on the predictions of a trained edge classifier. It assumes the input graph is almost noisy free (e.g., assortative graphs) such that an edge classifier can be trained reliably from the existing graph topology. However, this assumption does not hold for very noisy (disassortative) graphs that SGAT can handle. NeuralSparse~\cite{NeuralSparse20} learns a sparsification network to sample a $k$-neighbor subgraph (with a pre-defined $k$), which is then fed to GCN, GraphSage or GAT for node classification. Again, it does not aim to learn an edge-sparsified graph as the sparsification network produces a different subgraph sample each time and multiple subgraphs are used to improve accuracy. PTDNet~\cite{luo2021learning} proposes to improve the robustness and generalization performance of GNNs by learning to drop task-irrelevant edges. It samples a subgraph for each layer and applies a denoising layer before each GNN layer. Therefore, it cannot induce an edge-sparsified graph either. SuperGAT~\cite{SuperGAT21} improves GAT with an edge self-supervision regularization. It assumes that ideal attention should give all weights to label-agreed neighbors and introduces a layer-wise regularization term to guild attention with the presence or absence of an edge. However, when the graph is noisy, the regularization term will still push connected nodes to have same labels, and may generate suboptimal results. 

Overall, none of these prior works induce an edge-sparsified graph while retaining similar or improved classification accuracies. Moreover, all of these algorithms are evaluated on assortative graphs with improved performance. But none of them (except SuperGAT~\cite{SuperGAT21}) has been evaluated on noisy disassortative graphs. As we will see when we present results, SGAT outperforms all of these state-of-the-arts on disassortative graphs and demonstrates its robustness on assortative and disassortative graphs.



\section{Sparse Graph Attention Networks}\label{sec:method}
The key idea of our Sparse Graph Attention Networks (SGATs) is that we can attach a binary gate to each edge of a graph to determine if that edge shall be used for neighbor aggregation or not. We optimize the SGAT model under an $L_0$ regularized loss function such that we can use as fewer edges as possible to achieve similar or better classification accuracies. We first introduce our sparse attention mechanism, and then describe how the binary gates can be optimized via stochastic binary optimization.

\subsection{Formulation}\label{sec:formulation}
To identify important edges of a graph and remove noisy/task-irrelevant ones, we attach a binary gate $z_{ij}\in\{0,1\}$ to each edge $e_{ij}\in E$ such that $z_{ij}$ controls if edge $e_{ij}$ will be used for neighbor aggregation or not\footnote{Note that edges $e_{ij}$ and $e_{ji}$ are treated as two different edges and therefore have their own binary gates $z_{ij}$ and $z_{ji}$, respectively.}. This corresponds to attaching a set of binary masks to the adjacent matrix $A$:
\begin{equation}
  \bar{A}=A\odot Z, \quad\quad Z\in\{0,1\}^M,
\end{equation}
where $M$ is the number of edges in graph $G$. Since we want to use as fewer edges as possible for semi-supervised node classification, we train model parameters $W$ and binary masks $Z$ by minimizing the following $L_0$-norm regularized empirical risk:
\begin{align}\label{eq:l0}
  \mathcal{R}(W,Z) &\!=\!\frac{1}{n}\sum_{i=1}^{n}\mathcal{L}\left(f_i(X, A\odot Z, W), y_{i}\right) \!+\! \lambda\|Z\|_0 \\
  &\!=\!\frac{1}{n}\sum_{i=1}^{n}\mathcal{L}\left(f_i(X, A\odot Z, W), y_{i}\right) \!+\!\lambda\!\!\!\sum_{(i,j)\in E}\!\!\!1_{[z_{ij}\neq 0]}, \nonumber
\end{align}
where $\|Z\|_0$ denotes the $L_0$-norm of binary masks $Z$, i.e., the number of non-zero elements in $Z$ (edge sparsity), $1_{[c]}$ is an indicator function that is 1 if the condition $c$ is satisfied, and 0 otherwise, and $\lambda$ is a regularization hyperparameter that balances between data loss and edge sparsity. For the encoder function $f(X,A\odot Z,W)$, we define the following attention-based aggregation function:
\begin{equation}\label{eq:sgat_agg}
    h_{i}^{(l+1)} =\sigma\left(\sum_{j\in\mathcal{N}_i} a_{i j} h_{j}^{(l)}W^{(l)}\right), 
\end{equation}
where $a_{ij}$ is the attention coefficient assigned to edge $e_{ij}$ across all layers. This is in a stark contrast to GATs, in which a layer-dependent attention coefficient $a_{ij}^{(l)}$ is assigned for each edge $e_{ij}$ at layer $l$.

To compute attention coefficients, we simply calculate them by a row-wise normalization of $A\odot Z$, i.e.,
\begin{equation}\label{eq:sgat_attention}
  a_{i j}=\operatorname{normalize}\left(A_{ij}z_{i j}\right)=\frac{A_{ij}z_{i j}}{\sum_{k \in\mathcal{N}_{i}} A_{ik}z_{i k}}.
\end{equation}
Intuitively, the center node $i$ is important to itself; therefore we set $z_{ii}$ to 1 so that it can preserve its own information. Compared to GAT, we do not use softmax to normalize attention coefficients since by definition $z_{ij}\in\{0,1\}$ and typically $A_{ij}\geq 0$ such that their product $A_{ij}z_{ij}\geq 0$ .

Similar to GAT, we can also use multi-head attentions to increase the capacity of our model. We thus formulate a multi-head SGAT layer as:
\begin{equation}\label{eq:multi-head}
   h_i^{(l+1)}=\Big\|_{k=1}^{K}\sigma\left(\sum_{j \in\mathcal{N}_{i}} a_{i j} {h}_{j}^{(l)}W_k^{(l)}\right) ,
\end{equation}
where $K$ is the number of heads, $\|$ represents concatenation, $a_{ij}$ is the attention coefficients computed by Eq.~\ref{eq:sgat_attention}, and $W_k^{(l)}$ is the weight matrix of head $k$ at layer $l$. Note that only one set of attention coefficients $a_{ij}$ is calculated for edge $e_{ij}$, and they are shared among all heads and all layers. With multi-head attention, the final returned output, $h_i^{(l+1)}$, consists of $KD'$ features (rather than $D'$) for each node.

\begin{figure}[h]
	\begin{center}
		\includegraphics[width=1.0\linewidth]{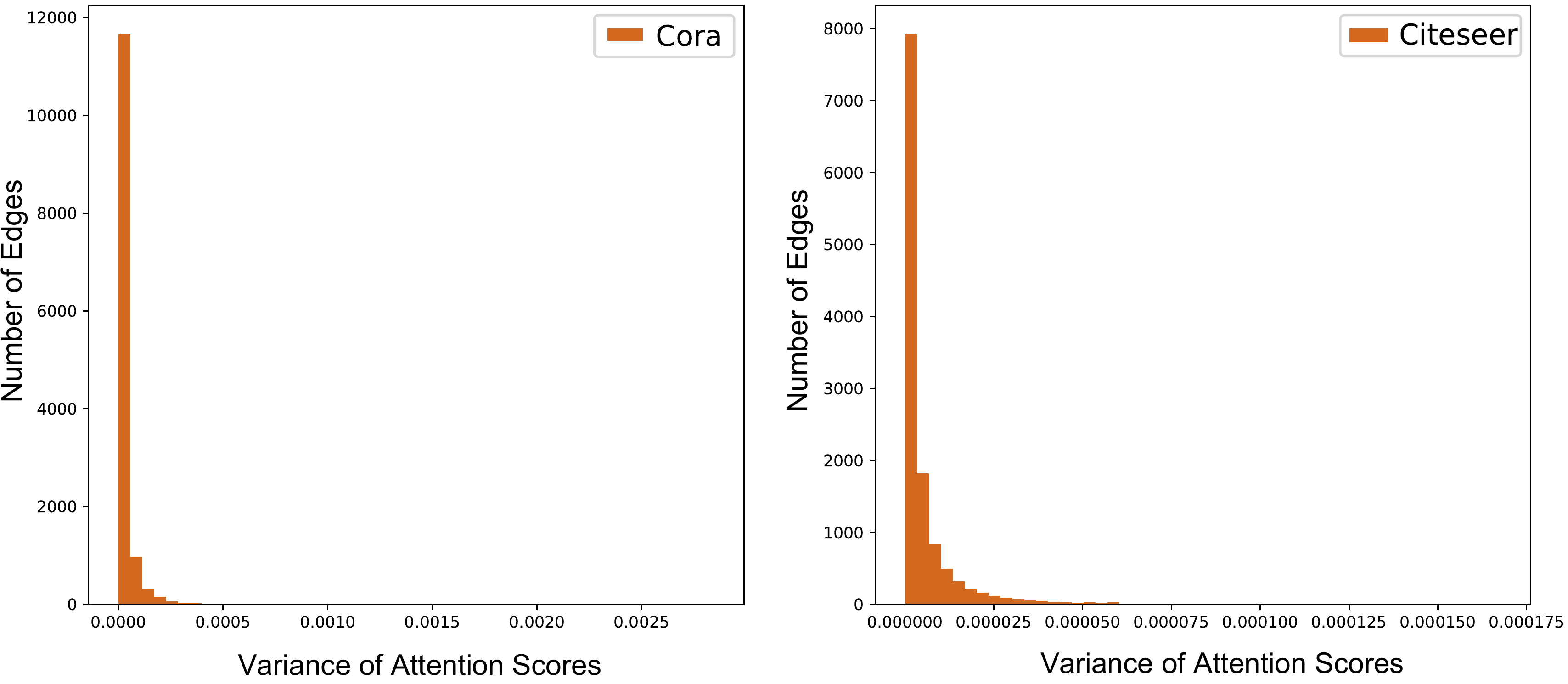}
	\end{center}\vspace{-10pt}
	\caption{Histogram of variance of attention coefficients of a 2-layer GAT with a 8-head attention on the Cora and Citeseer datasets. The variances of attention coefficients of majority of edges are close to 0, indicating GAT learns similar distributions of attention scores from all heads and all layers.}
	\label{fig:distribution}
\end{figure}

Why can we use one set of coefficients for multi-head attention? This is based on our observation that all GAT heads tend to learn attention coefficients with similar distributions, indicating significant redundancy in the GAT modeling. 
For example, given a 2-layer GAT with a 8-head attention, each edge receives 16 attention coefficients, on which the variance can be calculated. Fig.~\ref{fig:distribution} shows the histograms of variance of attention coefficients over all the edges in Cora and Citeseer, respectively\footnote{Similar patterns are observed on the other datasets used in our experiments.}. As we can see, the variances of attention coefficients of majority of edges are close to 0, indicating GAT learns similar distributions of attention coefficients from different heads and from different GAT layers. This means using one set of attention coefficients might be enough for feature aggregation. In addition, using one set of attention coefficients isn't rare in GNNs as GCNs use a shared $\hat{A}$ across all layers and are very competitive to GATs in terms of classification accuracies. While GCNs use one set of predefined aggregation coefficients, SGATs learn the coefficients from a sparse attention mechanism. We believe it is the learned attention coefficients instead of multi-set attention coefficients that leads to the improved performance of GATs over GCNs, and the benefit of multi-set attention coefficients might be very limited and could be undermined by the risk of overfitting due to increased complexity. Therefore, the benefits of using one set of attention coefficients over the original multi-set coefficients are at least twofold: (1) one set of coefficients is computationally $K$ times cheaper than multiple sets of coefficients and is less prone to overfitting; and (2) one set of coefficients can be interpreted as edge importance scores such that they can be used to identify important edges and remove noisy/task-irrelevant edges for robust learning from real-world graph-structured data.

\subsection{Model Optimization}

\textbf{Stochastic Variational Optimization}  
To optimize Eq.~\ref{eq:l0}, we need to compute its gradient w.r.t. binary masks $Z$. However, since $Z$ is a set of binary variables, neither the first term nor the second term is differentiable. Hence, we resort to approximation algorithms to solve this binary optimization problem. Specifically, we approximate Eq.~\ref{eq:l0} via an inequality from stochastic variational optimization~\cite{SVO18}: Given any function $\mathcal{F}(\boldsymbol{z})$ and any distribution $q(\boldsymbol{z})$, the following inequality holds:
\begin{equation}
   \min _{\boldsymbol{z}} \mathcal{F}(\boldsymbol{z}) \leq \mathbb{E}_{\boldsymbol{z} \sim q(\boldsymbol{z})}[\mathcal{F}(\boldsymbol{z})],
\end{equation}
i.e., the minimum of a function is upper bounded by its expectation. 

Since $z_{ij}\;\forall (i, j)\in E$ is a binary random variable, we assume $z_{i j}$ is subject to a Bernoulli distribution with parameter $\pi_{ij} \in[0,1],$ i.e. $z_{i j} \sim \operatorname{Ber}\left(z_{ij} ; \pi_{ij}\right)$. Thus, we can upper bound Eq.~\ref{eq:l0} by its expectation:
\begin{align}\label{eq:l0_exp}
   \tilde{\mathcal{R}}(W,\pi)\!=\!\frac{1}{n}\!\sum_{i=1}^{n}\!\mathbb{E}_{q\left(Z | \pi\right)} \mathcal{L}\!\left(f_i(X, A\odot Z, W), y_i\right)\!+\!\lambda\!\!\!\!\sum_{(i,j)\in E}\!\!\!\!\pi_{ij}.
\end{align}
Now the second term of Eq.~\ref{eq:l0_exp} is differentiable w.r.t. the new model parameters $\pi$. However, the first term is still problematic since the expectation over a large number of binary random variables $Z$ is intractable, and thus its gradient does not allow for an efficient computation. 

\vspace{10pt}
\noindent\textbf{The Hard Concrete Gradient Estimator} We therefore need further approximation to estimate the gradient of the first term of Eq.~\ref{eq:l0_exp} w.r.t. $\pi$. Fortunately, this is a well-studied problem in machine learning and statistics with many existing gradient estimators for this discrete latent variable model, such as REINFORCE~\cite{reinforce92}, Gumble-Softmax~\cite{gumbel-softmax17}, REBAR~\cite{rebar17}, RELAX~\cite{relax18} and the hard concrete estimator~\cite{l0sparse18}. We choose the hard concrete estimator due to its superior performance in our experiments and relatively straightforward implementation. Specifically, the hard concrete estimator employs a reparameterization trick to approximate the original optimization problem Eq.~\ref{eq:l0_exp} by a close surrogate function:
\begin{align} 
    \hat{\mathcal{R}}(\!W,\log\!\bs{\alpha}\!) \!&=\!\frac{1}{n}\!\sum_{i=1}^{n}\!\mathbb{E}_{\bs{u}\sim\mathcal{U}(0,1)}\mathcal{L}\Big(\!f_i(\!X, \!A\!\odot\!g(f(\log\!\bs{\alpha},\!\bs{u})),\!W),\!y_{i}\!\Big) \nonumber\\ &+\lambda \sum_{(i,j)\in E}\sigma\Big(\log \alpha_{ij}-\beta \log \frac{-\gamma}{\zeta}\Big) 
    \end{align}
 with
 \begin{align} 
 &f(\log\alpha, u)\!=\!\sigma((\log u\!-\!\log(1\!-\!u)\!+\!\log \alpha) / \beta)(\zeta\!-\!\gamma)\!+\!\gamma,\nonumber\\
 &g(\cdot)=\min (1, \max (0, \cdot)), \nonumber
 \end{align}
 where $\mathcal{U}(0,1)$ is a uniform distribution in the range of $[0,1]$, $\sigma(x)=\frac{1}{1+\exp(-x)}$ is the sigmoid function, and $\beta=2 / 3, \gamma=-0.1$   and $\zeta= 1.1$ are the typical parameter values of the hard concrete distribution. We refer the readers to~\cite{l0sparse18} for more details of the hard concrete gradient estimator.

During training, we optimize $\log\alpha_{ij}$ for each edge $e_{ij}$. At the test phrase, we generate a deterministic mask $\hat{Z}$ by employing the following formula:
\begin{equation}
    \hat{Z}=\min (\mathbf{1}, \max (\mathbf{0}, \sigma((\log \boldsymbol{\alpha})/\beta)(\zeta-\gamma)+\gamma)),
\end{equation}
which is the expectation of $Z$ under the hard concrete distribution $q(Z| \log \boldsymbol{\alpha})$. Due to the hard concrete approximation, $\hat{z}_{ij}$ is now a continuous value in the range of $[0,1]$. Ideally, majority of elements of $\hat{Z}$ will be zeros, and thus many edges can be removed from the graph.

\vspace{10pt}
\noindent\textbf{Inductive Model of $\log\bs{\alpha}$}
The learning of binary masks $Z$ discussed above is transductive, by which we can learn a binary mask $z_{ij}$ for each edge $e_{ij}$ in the training graph $G$. However, this approach cannot generate new masks for edges that are not in the training graph. A more desired approach is inductive that can be used to generate new masks for new edges. This inductive model of $\log\bs{\alpha}$ can be implemented as a generator, which takes feature vectors of a pair of nodes as input and produces a binary mask as output. We model this generator simply as
\begin{equation}\label{eq:asym}
  \log\alpha_{ij}=(x_iW_0^{(0)} \| x_jW_0^{(0)})b^T
\end{equation}
where $b\in \mathcal{R}^{D'}$ is the parameter of the generator and $W_0^{(0)}$ is the weight matrix of head 0 at layer 0. To integrate this generator into an end-to-end training pipeline, we define this generator to output $\log\alpha_{ij}$. Upon receiving $\log\alpha_{ij}$ from the mask generator, we can sample a mask $\hat{z}_{ij}$ from the hard concrete distribution $q(z|\log\alpha_{ij})$. The set of sampled mask $\hat{Z}$ is then used to generate an edge-sparsified graph for the downstream classification tasks. Fig.~\ref{fig:sgat} illustrates the full pipeline of SGATs. In our experiments, we use this inductive SGAT pipeline for semi-supervised node classification.

\begin{table*}[t]
\newcommand{\tabincell}[2]{\begin{tabular}{@{}#1@{}}#2\end{tabular}}
\caption{\label{dataset}Summary of the graph datasets used in the experiments}\vspace{-10pt}
\begin{center}
\begin{threeparttable}
\begin{tabular}{lllrrrrcc}
    \toprule
     & Type & Task &Nodes &Edges &Features &Classes  & \#Neighbors &$H(G)$ \\
    \midrule
    \tabincell{c}{\textbf{Cora}}   &assortative    &transductive  &2,708  & 13,264      &1,433     & 7   & 2.0    & 0.83  \\
    \tabincell{c}{\textbf{Citeseer}} &assortative       &transductive  &3,327  & 12,431      &3,703    & 6   & 1.4    & 0.71 \\
    \tabincell{c}{\textbf{Pubmed}}  &assortative        &transductive  &19,717 & 108,365     &500     & 3    & 2.3    & 0.79 \\
    \tabincell{c}{\textbf{Amazon computers}} &assortative &transductive  &13,381 & 505,474     &767     & 10   & 18.4   & 0.79 \\
  \tabincell{c}{\textbf{Amazon photo}}  &assortative    &transductive  &7,487  & 245,812     &745     & 8   & 15.9    & 0.84\\
  \midrule
  \tabincell{c}{\textbf{Actor}}   &disassortative          &transductive  &7,600    & 60,918      &931     & 5   & 4.4     & 0.24\\
  \tabincell{c}{\textbf{Cornell}}    &disassortative       &transductive  &183    & 737         &1,703    & 5   & 1.6     & 0.11\\
  \tabincell{c}{\textbf{Texas}}   &disassortative          &transductive  &183    & 741         &1,703    & 5   & 1.7     & 0.06\\
  \tabincell{c}{\textbf{Wisconsin}}  &disassortative       &transductive  &251    & 1,151        &1,703    & 5   & 2.0     & 0.16\\
	\midrule
    \tabincell{c}{\textbf{PPI }}    &assortative        &inductive     &56,944 & 818,716     &50      & 121 & 6.7     & -\tnote{*} \\
    \tabincell{c}{\textbf{Reddit }}   &assortative      &inductive     &232,965& 114,848,857  &602     & 41 & 246.0   & 0.81\\
    \bottomrule
\end{tabular}
    \begin{tablenotes}
      \item[*] PPI is a multi-label dataset, whose $H(G)$ can not be calculated.
  \end{tablenotes}
  \end{threeparttable}
\end{center}
\end{table*}

\section{Evaluation}

To demonstrate SGAT's ability of identifying important edges for feature aggregation, we conduct a series of experiments on synthetic and real-world (assortative and disassortative) semi-supervised node classification benchmarks, including transductive learning tasks and inductive learning tasks. We compare our SGATs with the state-of-the-art GNN algorithms: GCNs~\cite{kipf2017semi}, GraphSage~\cite{hamilton2017inductive}, GATs~\cite{velickovic2018graph}, SuperGAT~\cite{SuperGAT21}, DropEdge~\cite{dropedge} and PTDNet~\cite{luo2021learning}. For a fair comparison, our experiments closely follow the configurations of the competing algorithms. Our code is available at: \url{https://github.com/Yangyeeee/SGAT}.

\subsection{Graph Datasets}
\textbf{Assortative and Disassortative Graphs} Graph datasets can be categorized into assortative and disassortative ones~\cite{newman2002assortative,ribeiro2017struc2vec} according to the node homophily in terms of class labels as introduced by~\cite{pei2020geom},
\begin{equation}
  H(G) = \frac{1}{|V|}\!\sum_{v \in V}\!\frac{\text{Number of }v\text{’s neighbors of the same label  }}{\text{Number of }v\text{’s neighbors}}.\nonumber
\end{equation}
Assortative graphs refer to ones with a high node homophily, where nodes within the local neighborhood provide useful information for feature aggregation. Common assortative graphs include citation networks and community networks. On the other hand, disassortative graphs contain nodes of the same label but not in their direct neighborhood, and therefore nodes within the local neighborhood provide more noises than useful information. Example disassortative graph datasets are co-occurrence networks and webpage hyperlink networks. We evaluate our algorithm on both types of graphs to demonstrate the robustness of SGATs on pruning redundant edges from clean assortative graphs and noisy edges from disassortative graphs. In our experiments, we consider seven established assortative and four disassortative graphs, whose statistics are summarized in Table~\ref{dataset}. 

\vspace{5pt}
\noindent\textbf{Transductive Learning Tasks} Three citation network datasets: Cora, Citeseer and Pubmed~\cite{senaimag08} and two co-purchase graph datasets: Amazon Computers and Amazon Photo~\cite{pitfall19} are used to evaluate the performance of our algorithm in the transductive learning setting, where test graphs are \emph{included} in training graphs for feature aggregation and thus facilitates the learning of feature representations of test nodes for classification. The citation networks have low degrees (e.g., only 1-2 edges per node), while the co-purchase datasets have higher degrees (e.g., 15-18 edges per node). Therefore, we can demonstrate the performance of SGAT on sparse graphs and dense graphs. For the citation networks, nodes represent documents, edges denote citation relationship between two documents, and node features are the bag-of-words representations of document contents; the goal is to classify documents into different categories. For the co-purchase datasets, nodes represent products, edges indicate two products are frequently purchased together, and node features are the bag-of-words representations of product reviews; similarly, the goal is to classify products into different categories. Our experiments closely follow the transductive learning setup of ~\cite{kipf2017semi,pitfall19}. For all these datasets, 20 nodes per class are used for training, 500 nodes are used for validation, and 1000 nodes are used for test. 

For the four disassortative graphs, \textit{Actor}~\cite{tang2009social} is an actor co-occurrence network, where nodes denote actors and edges indicate co-occurrence of two actors on the same Wikipedia web page. Node features are the bag-of-word representation of keywords in the actors’ Wikipedia pages. Each node is labeled with one of five classes according to the topic of the actor's Wikipedia page. \textit{Cornell}, \textit{Texas}, and \textit{Wisconsin} come from the WebKB dataset collected by Carnegie Mellon University. Nodes represent web pages and edges denote hyperlinks between them. Node features are the bag-of-word representation of the corresponding web page. Each node is labeled with one of the five categories \{student, project, course, staff, and faculty\}. We follow~\cite{pei2020geom} to randomly split nodes of each class into 60\%, 20\%, and 20\% for training, validation, and test. The experiments are repeatedly run 10 times with different random splits and the average test accuracy over these 10 runs are reported. Test is performed when validation accuracy achieves maximum on each run.

\vspace{5pt}
\noindent\textbf{Inductive Learning Tasks} Two large-scale graph datasets: PPI~\cite{Zitnik2017} and Reddit~\cite{hamilton2017inductive} are also used to evaluate the performance of SGAT in the inductive learning setting, where test graphs are \emph{excluded} from training graphs for parameter learning, and the representations of test nodes have to be generated from trained aggregators for classification. In this case, our inductive experiments closely follow the setup of GraphSage~\cite{hamilton2017inductive}. The protein-protein interaction (PPI) dataset consists of graphs corresponding to different human tissues. Positional gene sets, motif gene sets and immunological signatures are extracted as node features and 121 gene ontology categories are used as class labels. There are in total 24 subgraphs in the PPI dataset with each subgraph containing 3k nodes and 100k edges on average. Among 24 subgraphs, 20 of them are used for training, 2 for validation and the rest of 2 for test. For the Reddit dataset, it's constructed from Reddit posts made in the month of September, 2014. Each node represents a reddit post and two nodes are connected when the same user commented on both posts. The node features are made up with the embedding of the post title, the average embedding of all the post's comments, the post's score and the number of comments made on the post. There are 41 different communities in the Reddit dataset corresponding to 41 categories. The task is to predict which community a post belongs to. We use the same data split as in GraphSage~\cite{hamilton2017inductive}, where the first 20 days for training and the remaining days for test (with 30\% used for validation). This is a large-scale graph learning benchmark that contains over 100 million edges and about 250 edges per node, and therefore a high edge redundancy is expected. 

\begin{figure*}[t]	
	\begin{center}
		\includegraphics[width=1\linewidth]{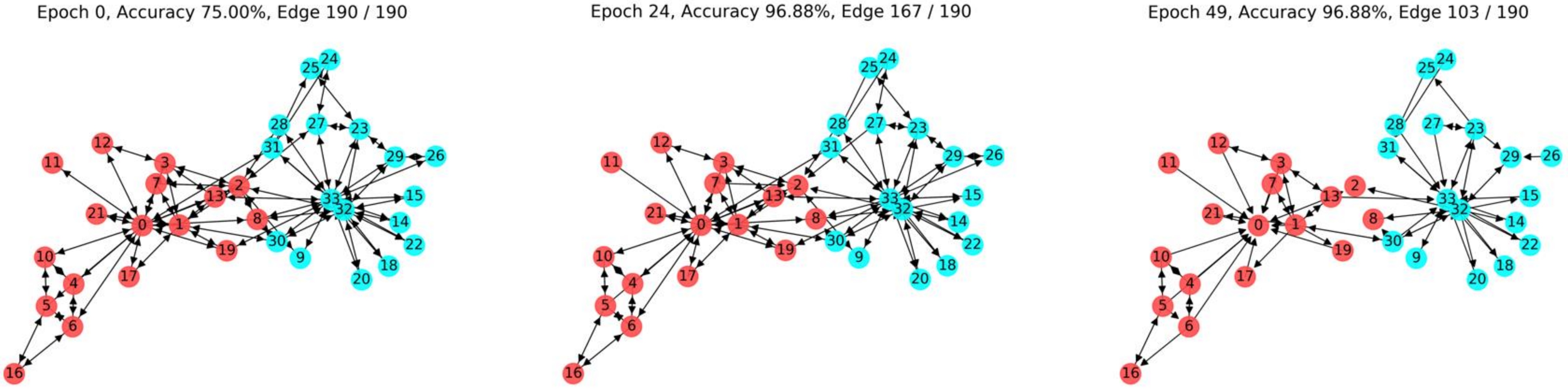}
	\end{center}\vspace{-0.6cm}
    \caption{The evolution of the graph of Zachary's Karate Club at different training epochs. SGAT can remove 46\% edges from the graph while retaining almost the same accuracy at 96.88\%. Nodes 0 and 33 are the labeled nodes, and the colors show the ground-truth labels. The video can be found at \url{https://youtu.be/3Jhr26lXRl8}.}
	\label{fig:demo}
\end{figure*}

\subsection{Models and Experimental Setup}
\textbf{Models} A 2-layer SGAT with a 2-head attention at each layer is used for feature aggregation, followed by a softmax classifier for node classification. We use ReLU~\cite{relu10} as the activation function and optimize the models with the Adam optimizer~\cite{adam2014} with the learning rate of $lr\!=\!1e\!-\!2$. We compare SGAT with the state-of-the-arts in terms of node classification accuracy. Since SGAT induces an edge-sparsified graph, we also report the percentage of edges removed from the original graph. 

To investigate the effectiveness of sparse attention mechanism induced by the $L_0$-norm regularization, we also introduce a GAT with \emph{top-k} attention baseline (named as GAT-2head-top-k). This baseline has the same architecture of SGAT-2head, which only uses one head's attentions to do neighbor aggregation. However, instead of using sparse attention coefficients induced by Eq.~\ref{eq:sgat_attention}, we remove the \emph{top-k} smallest attention coefficients calculated by GAT's dense attention function, with \emph{k} set to remove the same percentage of edges induced by SGAT-2head.  We report the best performance of GAT-2head-top-k from two training procedures: (1) removing the \emph{top-k} smallest attention coefficients from the beginning of the training, (2) removing \emph{top-k} smallest attention coefficients after the validation accuracies have converged. This GAT-2head-top-k baseline also produces an edge-sparsified graph, and thus serves as a fair comparison to SGAT-2head. 

\vspace{5pt}
\noindent\textbf{Hyperparameters} We tune the performance of SGAT and its variants based on the hyperparameters of GAT since SGAT is built on the basic framework of GAT. For a fair comparison, we also run 1-head and 2-head GAT models with the same architecture as SGATs to illustrate the impact of sparse attention models vs. standard dense attention models. To prevent models from overfitting on small datasets, $L_2$ regularization and dropout~\cite{dropout14} are used. Dropout is applied to the inputs of all layers and the attention coefficients. For the large-scale datasets, such as PPI and Reddit, we do not use $L_2$ regularization or dropout as the models have enough data for training. We implemented our SGAT and its variants with the DGL library~\cite{dgl19}. 


\subsection{Experiments on Synthetic Dataset}
To illustrate the idea of SGAT, we first demonstrate it on a synthetic dataset -- Zachary's Karate Club~\cite{zachary77}, which is a social network of a karate club of 34 members with links between pairs of members representing who interacted outside the club. The club was split into two groups later due to a conflict between the instructor and the administrator. The goal is to predict the groups that all members of the club joined after the split. This is a semi-supervised node classification problem in the sense that only two nodes: the instructor (node 0) and the administrator (node 33) are labeled and we need to predict the labels of all the other nodes. 

We train a 2-layer SGAT with a 2-head attention at each layer on the dataset. Fig.~\ref{fig:demo} illustrates the evolution of the graph at different training epochs, the corresponding classification accuracies and number of edges kept in the graph. As can be seen, as the training proceeds, some insignificant edges are removed and the graph is getting sparser; at the end of training, SGAT removes about 46\% edges while retaining an accuracy of 96.88\% (i.e., only one node is misclassified), which is the same accuracy achieved by GCN and other competing algorithms that utilize the full graph for prediction. In addition, the removed edges have an intuitive explanation. For example, the edge from node 16 to node 6 is removed while the reversed edge is kept. Apparently, this is because node 6 has 4 neighbors while node 16 has only 2 neighbors, and thus removing one edge between them doesn't affect node 6 too much while may be catastrophic to node 16. Similarly, the edges between node 27, 28 and node 2 are removed. This might be because node 2 has an edge to node 0 and has no edge to node 33, and therefore node 2 is more like to join node 0's group and apparently the edges to nodes 27 and 28 are insignificant or might be due to noise.

\subsection{Experiments on Assortative Graphs} 
Next we evaluate the performance of SGAT on seven assortative graphs, where nodes within the local neighborhood provide useful information for feature aggregation. In this case, some redundant or task-irrelevant edges may be removed from the graphs with no or minor accuracy losses. For a fair comparison, we run each experiments 10 times with different random weight initializations and report the average accuracies. 

\begin{table*}[t]
\newcommand{\tabincell}[2]{\begin{tabular}{@{}#1@{}}#2\end{tabular}}
\centering
\caption{\label{accuracy}Classification accuracies on seven assortative graphs for semi-supervised node classification. Results of GCNs on PPI and Reddit are trained in a transductive way. The results annotated with $^*$ are from our experiments, and the rest of results are from the corresponding papers. OOM indicates ``out of memory". Results are the averages of 10 runs.}
\begin{threeparttable}
\begin{tabular}{lccccccc}
    \toprule
     \textbf{Datasets}  & \textbf{Cora} & \textbf{Citeseer} & \textbf{Pubmed} & \begin{tabular}[c]{@{}l@{}}\textbf{Amazon} \\ \textbf{computer}\end{tabular}  & \begin{tabular}[c]{@{}l@{}}\textbf{Amazon} \\ \textbf{Photo}\end{tabular}  &\textbf{PPI} & \textbf{Reddit} \\
    \midrule
    \tabincell{c}{\textbf{GCN}~\cite{kipf2017semi}}                 & 81.5\%         & 70.3\%            & 79.0\%        &  81.5\%$^*$       & 91.2\%$^*$       & 50.9\%$^*$   & 94.38\%$^*$          \\
   \tabincell{c}{\textbf{GraphSage}~\cite{hamilton2017inductive}}            & -              & -                 & -             & -             & -            & 61.2\%        & 95.4\%              \\
	\tabincell{c}{\textbf{GAT}~\cite{velickovic2018graph}}$^*$               & 82.5\%         & 70.9\%            & 78.6\%        & 81.5\%        & 89.1\%       & 98.3\%        & OOM              \\
	\tabincell{c}{\textbf{GAT-1head}}$^*$         & 82.1\%         & 70.8\%            & 77.4\%        & 81.3\%        & 89.7\%       & 85.6\%        & 92.6\%              \\
	\tabincell{c}{\textbf{GAT-2head}}$^*$         & 83.5\%         & 70.8\%            & 78.3\%        & 82.4\%        & 90.4\%       & 97.6\%        & 93.5\%              \\
    \tabincell{c}{\textbf{GAT-2head-top-k}}$^*$ & 82.8\%         & 70.9\%            & 78.2\%        & 77.5\%        & 85.6\%       & 95.5\%        & 93.3\%              \\
    \tabincell{c}{\textbf{SGAT-1head}}$^*$      & 82.3\%         & 70.6\%            & 76.1\%        & 81.1\%        & 89.5\%       & 93.6\%        & 94.9\%             \\
	\tabincell{c}{\textbf{SGAT-2head }}$^*$       & 83.0\%         & 71.5\%            & 78.3\%        & 81.8\%        & 89.9\%       & 97.6\%        & 95.8\%              \\
	\midrule
    \tabincell{c}{\textbf{Edge Removed}}$^*$  &  2.0\%         & 1.2\%            & 2.2\%         & \textbf{63.6\%}& \textbf{42.3\%}& \textbf{49.3\%}& \textbf{80.8\%}  \\
    \bottomrule
\end{tabular}
\begin{tablenotes}
  \item $^*$From our experiments.
  \item Note that DropEdge~\cite{dropedge} and PTDNet~\cite{luo2021learning} and SuperGAT~\cite{SuperGAT21} have reported improved accuracies on assortative graphs. Hence, we do not include their results in this table since only SGAT induces edge-sparsified graphs and we only claim SGAT achieves similar accuracies as GAT on these graphs.
\end{tablenotes}
\end{threeparttable}
\end{table*}

The results are summarized in Table~\ref{accuracy}. Comparing SGAT with GCN, we note that SGAT outperforms GCN on the PPI dataset significantly while being similar on all the other six benchmarks. Comparing SGAT with GraphSage, SGAT again outperforms GraphSage on PPI by a significant margin. Comparing SGAT with GAT, we note that they achieve very competitive accuracies on all six benchmarks except Reddit, where the original GAT is ``out of memory" and SGAT can perform successfully due to its simplified architecture. Another advantage of SGAT over GAT is the regularization effect of the $L_0$-norm on the edges. To demonstrate this, we test two GAT variants: GAT-1head and GAT-2head that have the similar architectures as SGAT-1head and SGAT-2head but with different attention mechanisms (i.e., standard dense attention vs. sparse attention). As we can see, on the Reddit dataset, the sparse attention-based SGATs outperform GATs by 2-3\% while sparsifying the graph by 80.8\%. As discussed earlier, to evaluate the effectiveness of sparse attention mechanism of SGAT further, we also introduce a baseline (GAT-2head-top-k), which has the same architecture of SGAT-2head but removes the \emph{top-k} smallest coefficients calculated by GAT's dense attention function, with \emph{k} set to remove the same number of redundant edges induced by SGAT. As we can see from Table~\ref{accuracy}, SGAT-2head outperforms GAT-2head-top-k by 1\%-4\% accuracies on larger benchmarks (Amazon computer, Amazon Photo, PPI and Reddit) when a large percentage of edges is removed, demonstrating the superiority of SGAT's sparse attention mechanism over the naive top-k attention coefficients selection as used in GAT-top-k.

Overall, SGAT is very competitive against GCN, GraphSage and GAT in terms of classification accuracies, while being able to remove certain percentages of redundant edges from small and large assortative graphs. Specifically, on the three small citation networks: Cora, Citeseer and Pubmed, SGATs learn that majority of the edges are critical to maintain competitive accuracies as the original graphs are already very sparse (e.g., numbers of edges per node are 2.0, 1.4, 2.3, respectively. See Table~\ref{dataset}), and therefore SGATs remove only 1-2\% of edges. On the other hand, on the rest of large or dense assortative graphs, SGATs can identify significant amounts of redundancies in edges (e.g., 40-80\%), and removing them incurs no or minor accuracy losses.

\subsection{Experiments on Disassortative Graphs}
As shown in Table~\ref{dataset}, the $H(G)$ of disassortative graphs are around 0.1-0.2. This means the graphs are very noisy, i.e., a node and majority of its neighbors have different labels. In this case, the neighbor aggregation mechanism of GAT, GCN and GraphSage would aggregate noisy features from neighborhood and fail to learn good feature representations for the downstream classification tasks, while SGAT may excel due to its advantage of pruning noisy edges in order to achieve a high predictive performance. 

\begin{table}[t]
  \newcommand{\tabincell}[2]{\begin{tabular}{@{}#1@{}}#2\end{tabular}}
  \centering
  \caption{Classification accuracies of different node classification algorithms on four disassortative graphs. Results are the averages of 10 runs.}\label{result_dis}\vspace{-5pt}
  \begin{threeparttable}
  \begin{tabular}{lcccc}
      \toprule
      \textbf{Datasets} &\textbf{Actor} &\textbf{Cornell} &\textbf{Texas} &\textbf{Wisconsin} \\
      \midrule
          \tabincell{c}{\textbf{MLP}}$^*$ & 35.1 & 81.6 & 81.3 & 84.9  \\
          \tabincell{c}{\textbf{GAT}~\cite{velickovic2018graph}}$^*$ & 34.6 & 55.9 & 55.4 & 53.5 \\ 
           \tabincell{c}{\textbf{SuperGAT}~\cite{SuperGAT21}}$^*$ & 30.4 & 57.6 & 61.1 & 60.1 \\     
          \tabincell{c}{\textbf{DropEdge-GCN}~\cite{dropedge}}$^*$ & 30.6 & 54.5 & 61.5 & 59.8 \\
          \tabincell{c}{\textbf{Geom-GCN}~\cite{pei2020geom}} & 31.6 & 60.8 & 67.6 & 64.1 \\
          \tabincell{c}{\textbf{PTDNet-GCN}~\cite{luo2021learning}}$^*$ & 35.6 & 80.3 & 82.2 & 84.9 \\
      \tabincell{c}{\textbf{SGAT-2head}}$^*$   &$\textbf{35.7}$ & $\textbf{82.4}$ & $\textbf{86.2}$ & $\textbf{86.1}$ \\
      \midrule
      \tabincell{c}{\textbf{Edge Removed}}$^*$  & 88.1\%         & 93.9\%            & 95.0\%         & 91.9\%  \\
      \bottomrule
  \end{tabular}
    \begin{tablenotes}
      \item $^*$From our experiments.
    \end{tablenotes}
  \end{threeparttable}
\end{table}

To verify this, we compare SGAT with GAT, Geom-GCN~\cite{pei2020geom}, MLP, DropEdge, SuperGAT and PTDNet on the four disassortative graphs. Geom-GCN is a variant of GCN that utilizes a complicated node embedding method to identify similar nodes and create an edge between them, such that it can aggregate features from informative nodes and outperform GCN on the disassortative graphs. We also consider an MLP model as baseline, which makes prediction solely based on the node features without aggregating any local information. For a fair comparison, the GAT and MLP have a similar model capacity as that of SGAT-2head. We also compare SGAT with the state-of-the-art robust GNN models that we discussed in related work: SuperGAT~\cite{SuperGAT21}\footnote{https://github.com/dongkwan-kim/SuperGAT}, DropEdge~\cite{dropedge}\footnote{https://github.com/DropEdge/DropEdge}, and PTDNet~\cite{luo2021learning}\footnote{https://github.com/flyingdoog/PTDNet}. Since none of them reported the performance on these disassortative graphs, we follow the same experimental settings discussed above and run their open source implementations.

The results are shown in Table~\ref{result_dis}. It can be observed that MLP outperforms GAT and the majority of algorithms considered, indicating that local aggregation methods fail to get a good performance due to the noisy neighbors. The robust GNN algorithms: SuperGAT and DropEdge achieve better performance than GAT in general but are still worse than MLP since the extremely noisy neighbors violate the label-agreement assumption of SuperGAT or beyond the noise level that simple drop edge can handle. On the other hand, PTDNet achieves a competitive performance with MLP, demonstrating that PTDNet's denoising layers and layer-wise subgraph sampling are indeed very effective. Among all the algorithms considered, SGAT achieves the best accuracies on the disassortative graphs. As shown in the last row of Table~\ref{result_dis}, on all the dissassortative graphs SGAT tends to remove majority of edges from the graphs, and only less than 10\% edges are kept for feature aggregation, which explains its superior performance on these noisy disassortative graphs.  

Overall, SGAT is a much more robust algorithm than GAT (and in many cases other competing methods) on assortative and disassortative graphs since it can detect and remove noisy/task-irrelevant edges from graphs in order to achieve similar or improved accuracies on the downstream classification tasks.


\subsection{Analysis of Removed Edges} 
\begin{figure}[t]
	\begin{center}
		\includegraphics[width=1.0\linewidth]{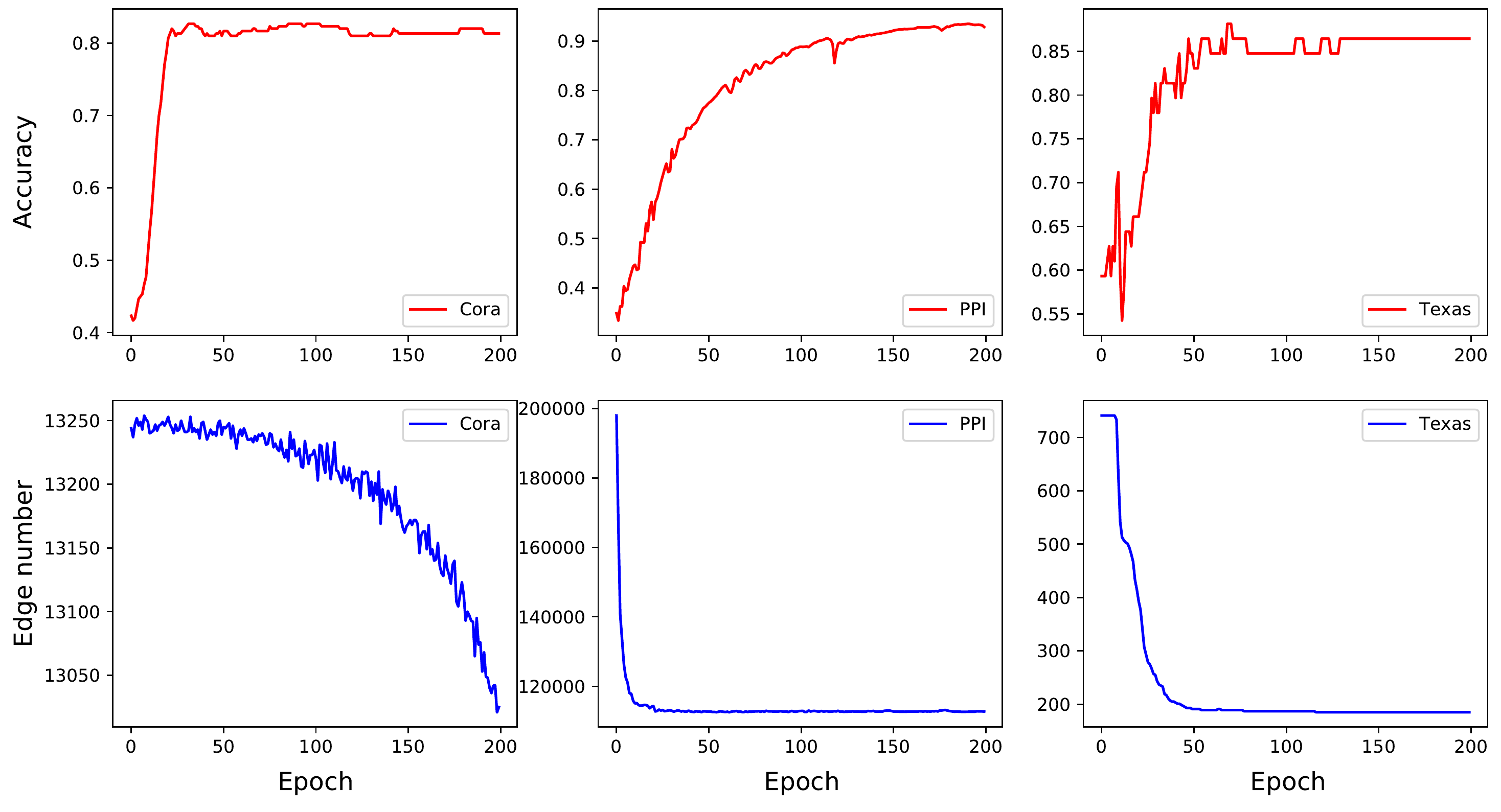}
	\end{center}\vspace{-15pt}
	\caption{The evolution of classification accuracy (top) and number of kept edges (bottom) as a function of training epochs on the Cora, PPI and Texas test datasets.}
	\label{fig:acc}
\end{figure}

We further analyze the edges removed by SGAT. Fig.~\ref{fig:acc} illustrate the evolution of classification accuracy and number of edges kept by SGAT as a function of training epochs on the Cora, PPI and Texas test datasets. As we can see, SGAT removes 2\% edges from Cora slowly during training (as Cora is a sparse graph), while it removes 49.3\% edges from PPI and over 88.1\% edges from Texas rapidly, indicating a significant edge redundancy in PPI and Texas.

\begin{figure}[h]	
	\begin{center}
		\includegraphics[width=0.9\linewidth]{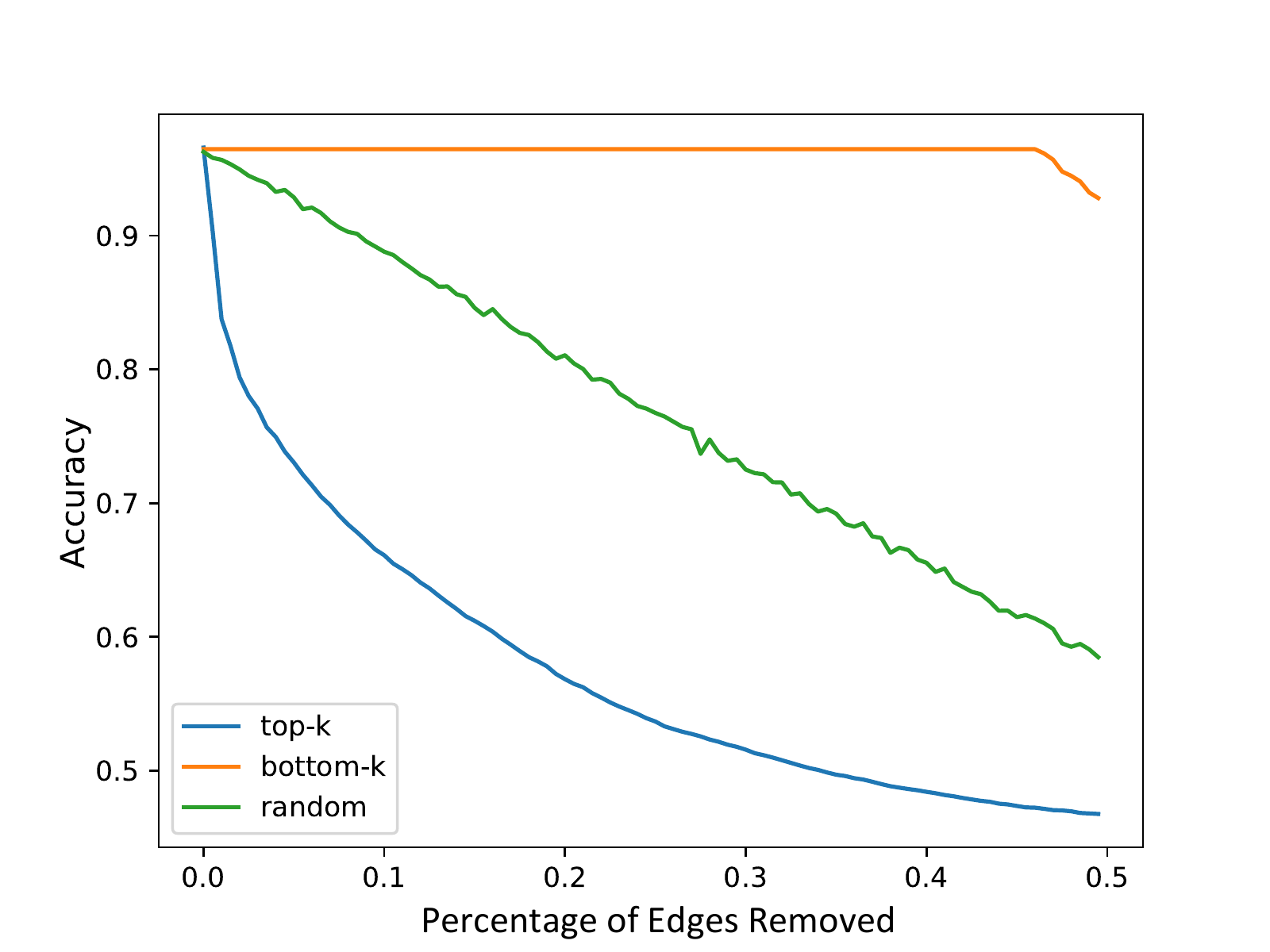}
	\end{center}\vspace{-15pt}
	\caption{The evolution of classification accuracies on the PPI test dataset when different percentages of edges are removed from the graph. Three different strategies of selecting edges for removal are considered.}
	\label{fig:topk}
\end{figure}

To demonstrate SGAT's accuracy of identifying important edges from a graph, Fig.~\ref{fig:topk} shows the evolution of classification accuracies on the PPI test dataset when different percentages of edges are removed from the graph. We compare three different strategies of selecting edges for removal: (1) top-k\% edges sorted descending by $\log\alpha_{ij}$, (2) bottom-k\% edges sorted descending by $\log\alpha_{ij}$, and (3) uniformly random k\%. As we can see, SGAT identifies important edges accurately as removing them from the graph incurs a dramatically accuracy loss as compared to random edge removal or bottom-k\% edge removal. 

\subsection{Hyperparameter Tuning}
\begin{figure}[t]	
	\begin{center}
		\includegraphics[width=0.9\linewidth]{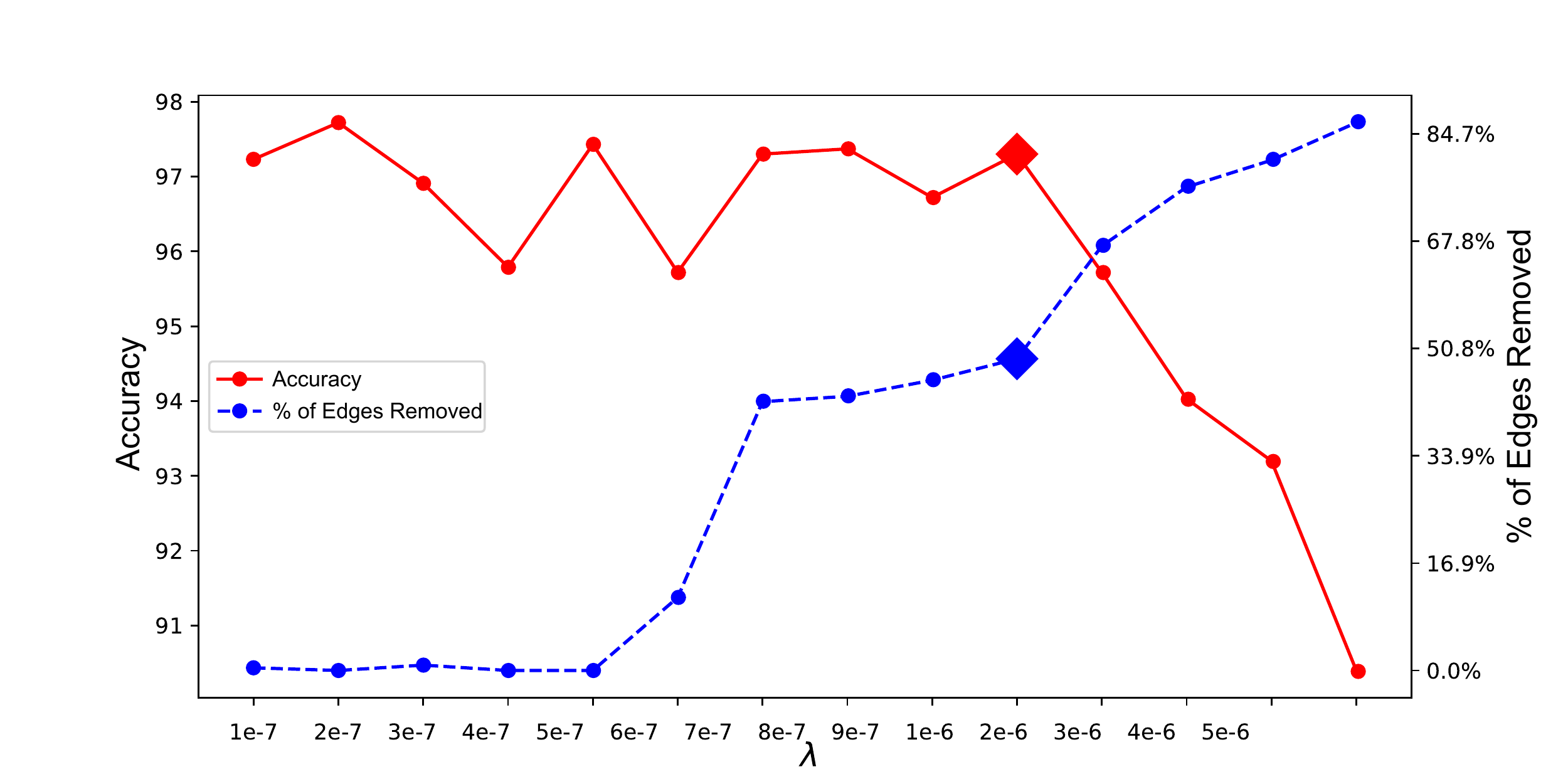}
    	\includegraphics[width=0.9\linewidth]{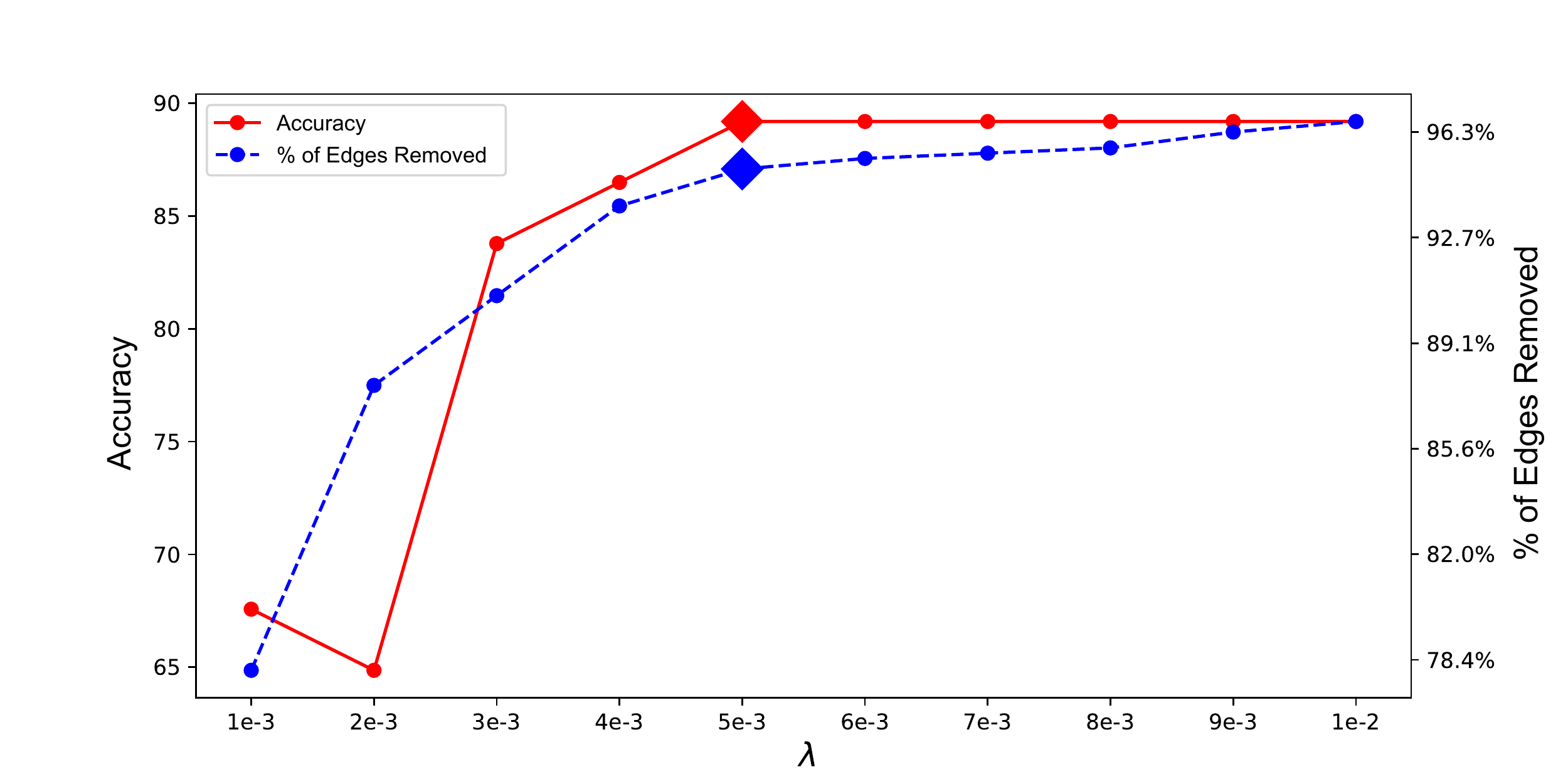}
	\end{center}\vspace{-15pt}
	\caption{The impact of $\lambda$ to the classification accuracy and edge sparsity on the PPI (top) and Texas (bottom) validation dataset.}
	\label{fig:lambda}
\end{figure}

SGAT has a few important hyperparameters, which affect the performance of SGAT significantly. In this section, we demonstrate the impact of them and discuss how we tune the hyperparameters for performance trade-off. One of the most important hyperparameters of SGAT is the $\lambda$ in Eq.~\ref{eq:l0}, which balances the classification loss (the first term) and edge sparsity  (the second term). As $\lambda$ increases, the $L_0$ sparsity regularization gets stronger. As a result, a large number of edges will be pruned away (i.e., $z$=0), but potentially it will incur a lower classification accuracy if informative edges are removed (i.e., over-pruning). We therefore select a $\lambda$ to yield the highest edge prune rate, while still achieving a good predictive performance on the downstream classification tasks. Fig.~\ref{fig:lambda} shows the results of tuning $\lambda$ on the PPI (top) and Texas (bottom) validation datasets. It can be observed that as $\lambda$ increases, more edges are removed from the PPI and Texas datasets. However, the classification accuracies have different trends on PPI and Texas. As more edges are removed from PPI, the accuracy retains almost no changes when $\lambda\!\le\!2e\!-\!6$, and drops significantly when $\lambda\!>\!2e\!-\!6$. This is because PPI is an assortative graph, in which local neighborhood provides useful information for feature aggregation, and pruning them from the graph in general incurs no or minor accuracy loss until a large $\lambda$ that leads to over-pruning. In contrast, as more edges are removed from Texas, the accuracy increases at beginning when $\lambda\!\le\!5e\!-\!3$ and plateaus afterwards. This is because Texas is a disassortative graph, in which local neighborhood provides more noise than useful information for feature aggregation, and pruning noisy edges from the graph typically improves classification accuracy until a large $\lambda$ that leads to over-pruning. Similar patterns are observed on the other assortative and disassortative graphs used in our experiments. Based on the results in Fig.~\ref{fig:lambda}, we choose $\lambda\!=\!2e\!-\!6$ for PPI and $\lambda\!=\!5e\!-\!3$ for Texas as they achieve the best balance between classification accuracy and edge sparsity.

\begin{figure}[h]
	\begin{center}
		\includegraphics[height=0.45\linewidth, width=1.0\linewidth]{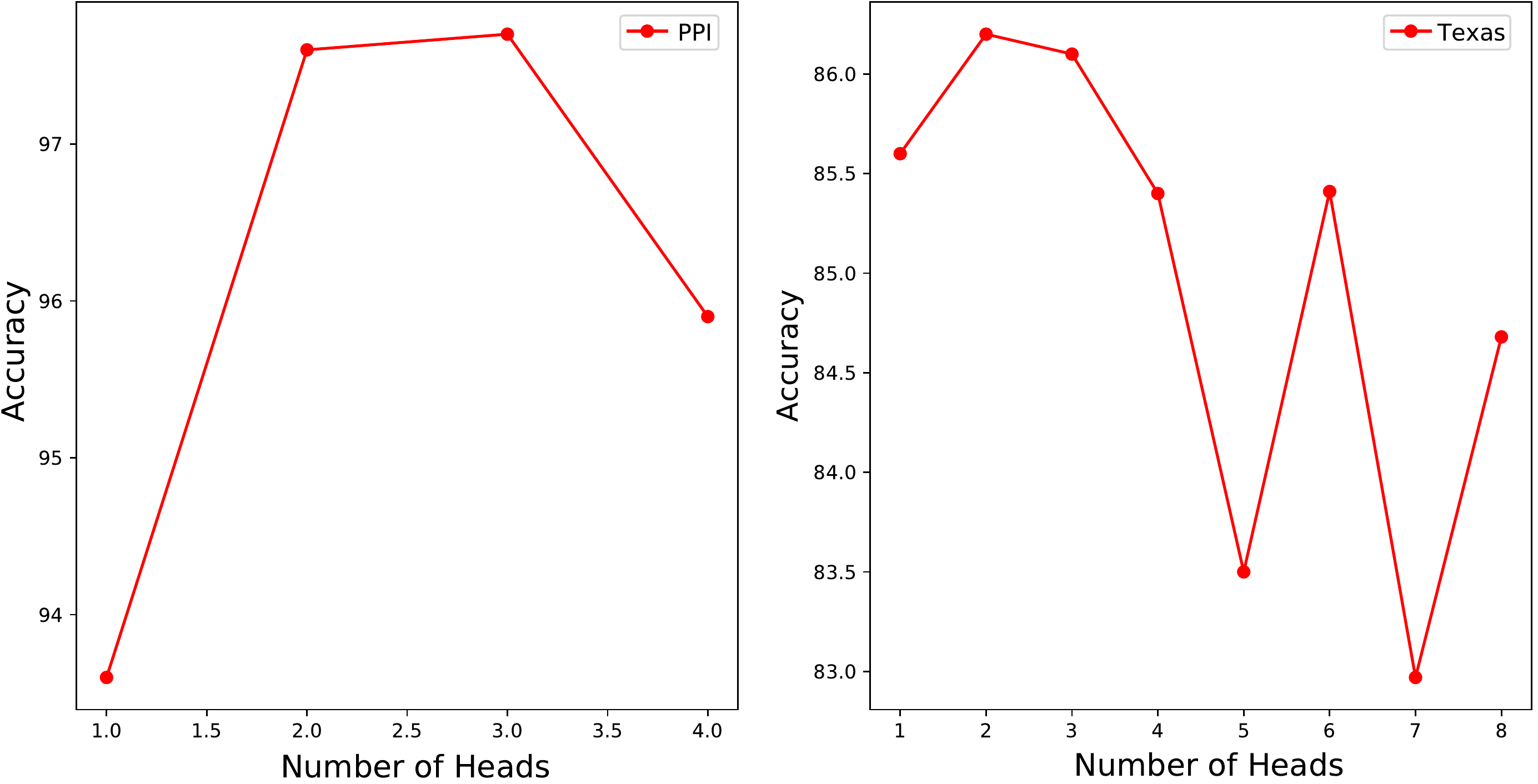}
	\end{center}\vspace{-15pt}
	\caption{The evolution of classification accuracy of SGAT as a function of number of heads on the PPI and Texas datasets.}
	\label{fig:head}
\end{figure}

Another important hyperparameter of SGAT is the number of heads $K$ in Eq.~\ref{eq:multi-head}. As $K$ increases, SGAT has more capacity to learn from the data, but is more prone to overfitting. This is demonstrated in Fig.~\ref{fig:head}, where we present the classification accuracies of SGAT on PPI and Texas as $K$ increases. As we can see, when $K=2$ SGAT achieves the best (or close to best) accuracies on both datasets. Similar trends are also observed on the other datasets. Therefore, in our experiments we choose $K\!=\!2$ as the default.

\subsection{Visualization of Learned Features}
\begin{figure}[h]	
	\begin{center}
		\includegraphics[width=1\linewidth]{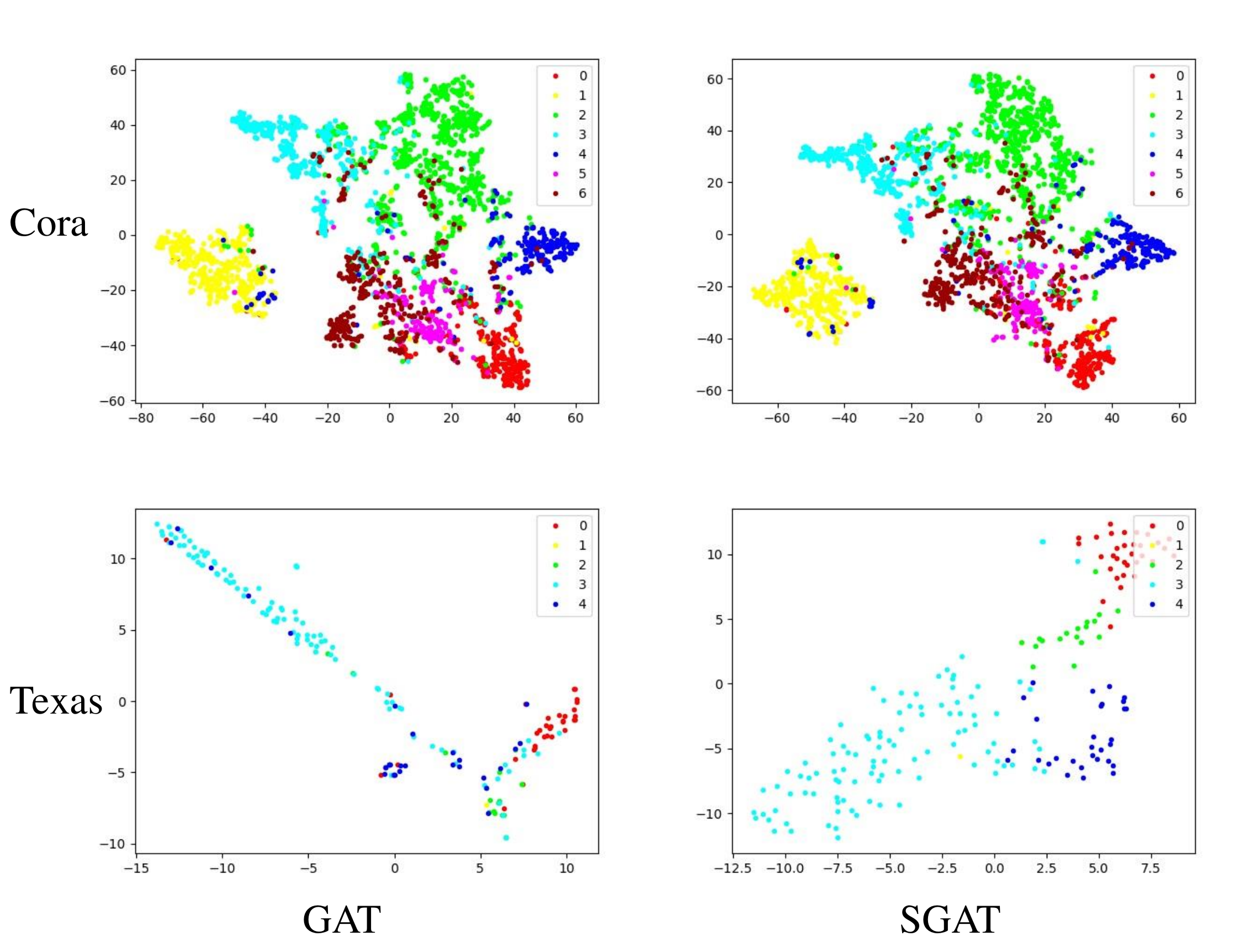}
	\end{center}\vspace{-10pt}
	\caption{t-SNE visualization of learned feature representations on Cora and Texas.}
	\label{fig:tsne}
\end{figure}

Finally, we visualize the learned feature representations from the penultimate layer\footnote{The layer before the final FC  layer for classification.} of GAT and SGAT with t-SNE~\cite{tsne}. The results on Cora and Texas are shown in Fig.~\ref{fig:tsne}. It can be observed that SGAT and GAT learn similar representations on Cora when the graph is nearly noisy-free (e.g., assortative graphs), while SGAT learns a better representation with a higher class separability than GAT on Texas when the graph is very noisy (e.g., disassortative graphs), demonstrating the robustness of SGAT on learning from assortative and disassortative graphs.

\subsection{Discussion}
Given a similar architecture and the same number of heads, one may expect that SGAT would be faster and more memory efficient than GAT since a large portion of edges can be removed by SGAT. However, our empirical study shows that both algorithms have a similar overall runtime and memory consumption. This is because learning sparse attention coefficients has the similar complexity as learning standard dense attention coefficients and storing feature representations (other than $A$ and $Z$) consumes most of memory. Therefore, even though SGAT can remove a large portion of edges from a graph, it isn't faster or more memory efficient than GAT. 

One potential speed up of SGAT is that we can skip the computation associated with edges of $z\approx 0$ during training. However, this heuristic will be an approximation because an edge with $z\approx 0$ may be reactivated in later iterations during stochastic optimization, and this potentially will cause accuracy drop. We leave this as our future work.

In summary, the main advantage of SGAT is that it can identify noisy/task-irrelevant edges from both assortative and disassortative graphs to achieve a similar or improve classification accuracy, while the conventional GAT, GCN and GraphSage fail on noisy disassortative graphs due to their local aggregation mechanism. The robustness of SGAT is of practical importance as real-world graph-structured data are often very noisy, and a robust graph learning algorithm that can learn from both assortative and disassortative graphs is very critical.

\section{Conclusion}
In this paper we propose sparse graph attention networks (SGATs) that integrate a sparse attention mechanism into graph attention networks (GATs) via an $L_0$-norm regularization on the number of edges of a graph. To assign a single attention coefficient to each edge, SGATs further simplify the architecture of GATs by sharing one set of attention coefficients across all heads and all layers. This results in a robust graph learning algorithm that can detect and remove noisy/task-irrelevant edges from a graph in order to achieve a similar or improved accuracy on downstream classification tasks. Extensive experiments on seven assortative graphs and four disassortative graphs demonstrate the robustness of SGAT.



As for future extensions, we plan to investigate the applications of SGATs on detecting superficial or malicious edges injected by adversaries. We also plan to explore the application of sparse attention network of SGATs in unsupervised graph domain adaption (e.g.~\cite{UDAGCN20}) to improve inter-graph attention.

\section{Acknowledgment}
We would like to thank the anonymous reviewers for their comments and suggestions, which helped improve the quality of this paper.

\ifCLASSOPTIONcaptionsoff
  \newpage
\fi



%



\bibliographystyle{IEEEtran}
\bibliography{IEEEabrv,gcn}

\end{document}